\documentclass[10pt,journal,letterpaper,compsoc]{IEEEtran}

\usepackage[utf8]{inputenc} 
\usepackage[T1]{fontenc}    
\usepackage{hyperref}       
\usepackage{url}            
\usepackage{booktabs}       
\usepackage{amsfonts}       
\usepackage{nicefrac}       
\usepackage{microtype}      
\usepackage[square,sort,comma,numbers]{natbib}
\usepackage{graphicx}
\usepackage{multirow}
\usepackage{subfigure}
\usepackage[ruled]{algorithm2e}
\usepackage{epsfig}
\usepackage{color}
\usepackage{amsthm}
 \usepackage{amsmath}
\usepackage{amssymb}

\usepackage{footmisc}
\usepackage{bm}
\usepackage{epstopdf}

\def\V05{\hspace{-0.075in}}
\def\VV05{\vspace{-0.01in}}

\title{DcnnGrasp: Towards Accurate Grasp Pattern Recognition with Adaptive Regularizer Learning}

\author{Xiaoqin Zhang, Ziwei Huang, Jingjing Zheng
, Shuo Wang, and Xianta Jiang

\IEEEcompsocitemizethanks{
\IEEEcompsocthanksitem
X. Zhang and Z. Huang are with the College of Computer Science and Artificial Intelligence, Wenzhou University, Zhejiang, China (e-mail:
		zhangxiaoqinnan@gmail.com, liyucheng0318@gmail.com).

\IEEEcompsocthanksitem
J. Zheng, S. Wang and X. Jiang are with the Department of Computer Science, Memorial University of Newfoundland, Newfoundland and Labrador, Canada (e-mail: jjzheng233@gmail.com, shuow@mun.ca, xiantaj@mun.ca).
}
}

\begin{document}

\IEEEcompsoctitleabstractindextext{
	\begin{abstract}
	The task of grasp pattern recognition aims to derive the applicable grasp types of an object according to the visual information. Current state-of-the-art methods ignore category information of objects which is crucial for grasp pattern recognition. This paper presents a novel dual-branch convolutional neural network (DcnnGrasp) to achieve joint learning of object category classification and grasp pattern recognition. DcnnGrasp takes object category classification as an auxiliary task to improve the effectiveness of grasp pattern recognition. Meanwhile, a new loss function called joint cross-entropy with an adaptive regularizer is derived through maximizing a posterior, which significantly improves the model performance. Besides, based on the new loss function, a training strategy is proposed to maximize the collaborative learning of the two tasks. The experiment was performed on five household objects datasets including the RGB-D Object dataset, Hit-GPRec dataset, Amsterdam library of object images (ALOI), Columbia University Image Library (COIL-100), and MeganePro dataset 1. The experimental results demonstrated that the proposed method can achieve competitive performance on grasp pattern recognition with several state-of-the-art methods. Specifically, our method even outperformed the second-best one by nearly 15\% in terms of global accuracy for the case of testing a novel object on the RGB-D Object dataset.

	\end{abstract}
	
	}

    \maketitle

	\section{Introduction}\label{section:Introduction}
\begin{figure}[t]
	\centering
        	\subfigure[Deep learning-based artificial vision system for robot hand control]{
		\includegraphics[scale=0.2]{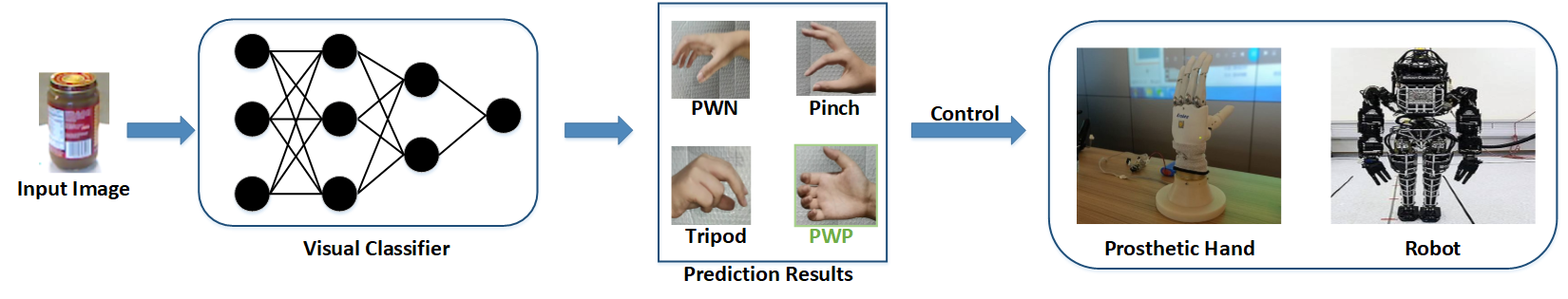}}
			\subfigure[The unseen problem in Grasp Pattern Recognition]{
		\includegraphics[scale=0.3]{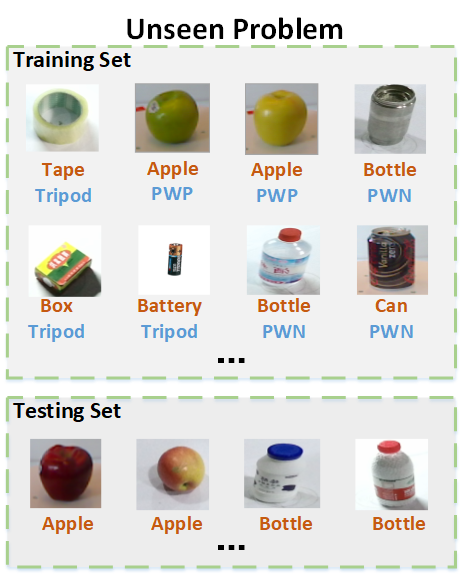}}
		\quad
			\subfigure[Challenges in Grasp Pattern Recognition]{
		\includegraphics[scale=0.28]{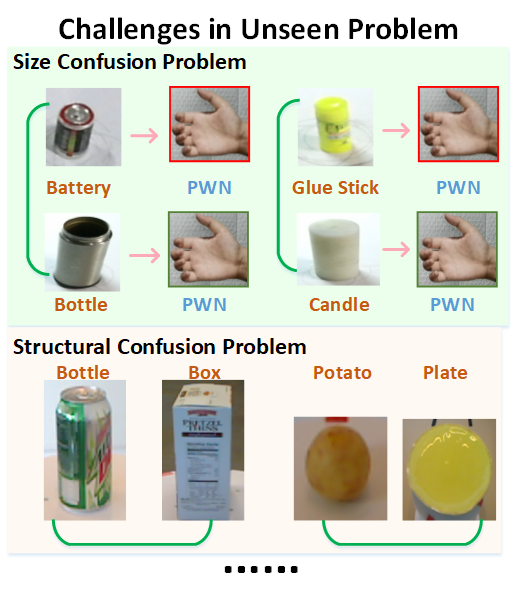}}
		\caption{(a) Deep learning-based artificial vision system for robot hand control: the visual classifier predicts the grasp type for household objects in the input image and then controls the device to perform the corresponding grasping action, where PWN and PWP refer to palmar wrist neutral and palmar wrist pronated, respectively. (b) The unseen problem in grasp pattern recognition, where the objects in the testing images have never appeared in the training set. (c) Challenges in grasp pattern recognition (especially in the unseen problem), such as the size confusion problem and structural confusion problem.}\label{problems}
	\end{figure}

    The goal of computer vision-based grasp pattern recognition is to derive the grasp type according to visual information of household objects, thus facilitating the grasping process of upper limb prosthesis control or gesture-based human robot interaction. A typical grasping process consists of grasp pattern recognition based on a visual classifier and robot hand control, as shown in Fig.~\ref{problems} (a). In the past years, many grasp pattern recognition methods have been proposed, which have a great impact in several fields including robot imitation \cite{wang2017robust,liu2018imitation,di2020safari}, human-robot interaction \cite{jin2020smart,wang2021smarthand},  human grasp predicting \cite{corona2020ganhand,zandigohar2021multimodal,ghazaei2019grasp,zandigohar2021netcut,veres2017modeling}, prosthetic design \cite{weiner2018kit,liu2021novel,yusof2021design}, robot control \cite{kumra2017robotic,du2019vision,caldera2018review,song2020novel,deng2019attention}, and prosthetic control \cite{taverne2019video,hundhausen2019resource,kara2019modeling}.
    	In recent years, Convolutional Neural Network (CNN) \cite{gu2018recent,krizhevsky2012imagenet} has shown excellent performance on many challenging tasks, such as object detection and tracking. It provides a new solution for grasp pattern recognition.
		As a pioneering work of the CNN-based method, \cite{RN2} proposed a CNN with a simple structure for grasp pattern recognition (it is called CnnGrasp in this paper). In \cite{RN2}, Ghazaei  \emph{et al.} labeled the images of household objects into corresponding grasp types and converted grasp pattern recognition to an image classification problem.
		  Inspired by this work, a series of CNN-based grasp pattern recognition methods emerged \cite{hundhausen2019resource,ghazaei2019grasp}.

	
    However, grasp pattern recognition has several significant problems. One is the presence of unseen objects during evaluation, in which a tested object and its views are wholly unseen by algorithms during training, such as the example given in Fig.~\ref{problems} (b) \cite{RN2}. In the bottom panel, the tested object (red apple) has never appeared in the training set. In this case, it is usually difficult for the traditional model to predict correctly.



    The unseen problems such as the size confusion problem and the structural confusion problem can affect the model performance greatly. As shown in Fig.~\ref{problems} (c), the battery and the glue stick are mis-recognized when `PWN' instead of `Tripod' is used due to their `big sizes' shown in the images. The structural confusion problem occurs when different objects are shot at certain angles, which leads to similar appearances of the objects in the images, such as the bottle and box as shown in Fig.~\ref{problems} (c). It can be seen that the image taken from the bottle is confused as a rectangle, and the potato may be confused as a plate. Therefore, the grasp types may be mis-recognized in these cases.


In addition, compared with object categories labeling, grasp type labeling is much more complex and takes a lot of human resources, considering the varying hand controlling requirements and gesture annotations of different applications and tasks. For example, in some tasks, a higher grasping precision is required, and the four commonly used gestures in daily life are no longer sufficient. Another example is prosthetic control, in which the influence of personal habits on grasping gestures should be considered. These factors all affect the gesture annotation results. Also, it is labor-consuming and time-costing to re-label gesture data to adapt to different applications and scenarios. Therefore, the following problem arises: how to train a relatively powerful grasp classifier with only a few grasp labeling results that are given in advance.

One simplest idea for handling this problem is to introduce object categories information. For the human being, even if part of the visual information of the object is missing or the objects are unseen before, one can still figure out the corresponding grasp type according to their experience of using the same category of objects before. This is because object category information usually contains the necessary information including geometric structure, size, and even material of the objects. However, this information is ignored by the existing grasp pattern recognition methods. Thus, it is a natural thought to jointly learn object category classification and grasp pattern recognition to improve grasp pattern recognition.

    Motivated by the above observation, this paper proposes a new deep learning architecture called Dual-branches Convolutional Neural Network (DcnnGrasp) for effective grasp pattern recognition. DcnnGrasp has two branches, namely the object category classification branch and the grasp pattern recognition branch. The former can help to achieve better grasp pattern recognition precision. Since the recognition tasks share a strong correlation, it is natural to jointly learn object category information and object grasp type information. To our best knowledge, this is the first work that utilizes object category information.

\begin{figure*}[t]
		\centering
        \includegraphics[scale=0.3]{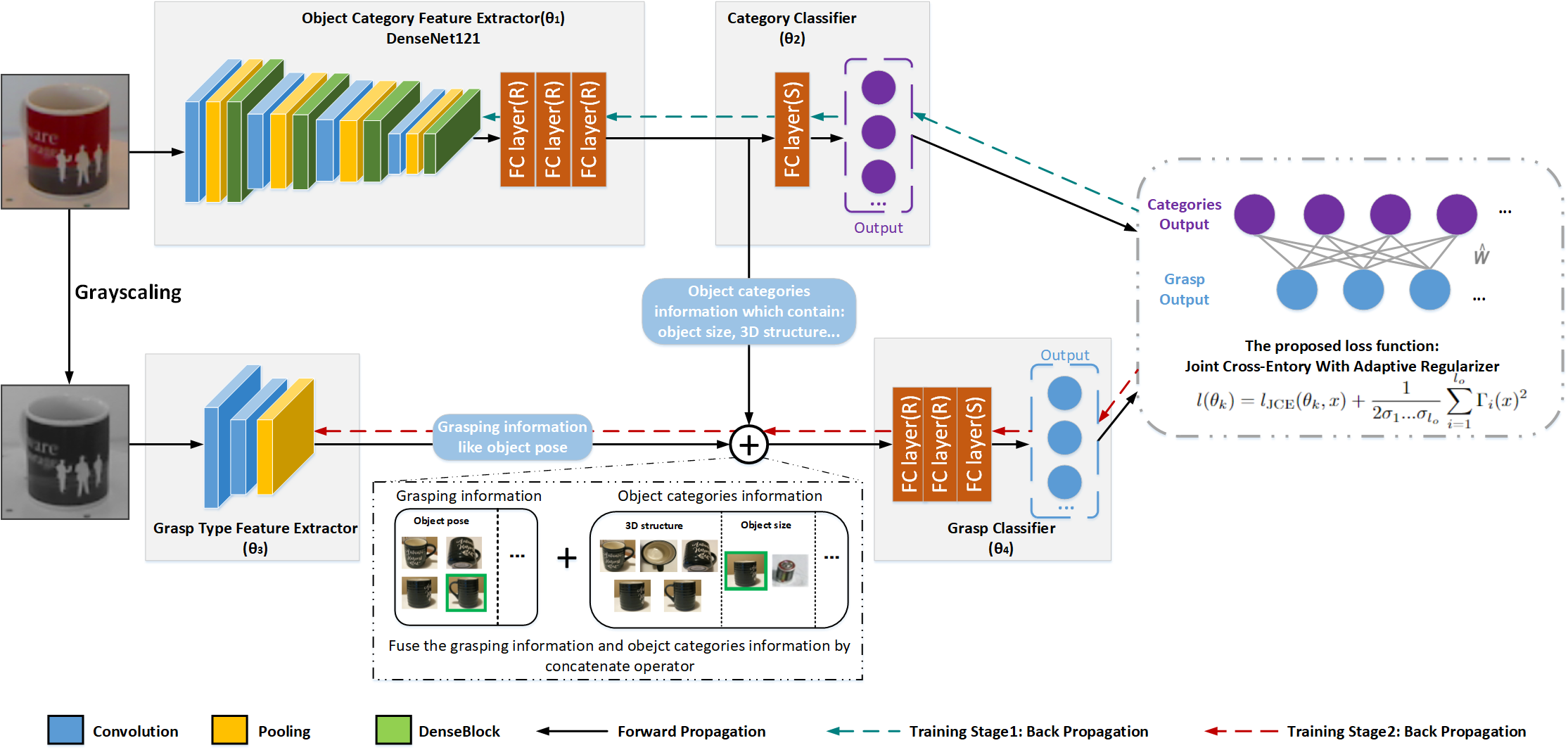}
		\caption{The architecture of DcnnGrasp. The model consists of four parts, object category feature extractor, category classifier, grasp type feature extractor, and grasp classifier. The newly proposed loss function called Joint Cross-Entropy is combined with Adaptive Regularizer (JCEAR) to train the network, where $\hat{W}$ is a conditional probability matrix obtained by statistical methods on the training set. The matrix $\hat{W}$ reflects the relationship between object categories and grasp types. To enable the model to achieve better performance, this paper proposes a training strategy based on (JCEAR). Backpropagation stage 1 and stage 2 come from our proposed training strategy.}\label{new_overview-network}
	\end{figure*}

    The contributions of this work are three-folds:
		\begin{itemize}
		\item A novel dual-branch network structure (DcnnGrasp) is proposed for grasp pattern recognition (See Fig.~\ref{new_overview-network}), in which the features extracted from the object category branch and the grasp type branch are integrated to enhance the performance of grasp pattern recognition.
		To train DcnnGrasp,  four dual-label datasets with grasp types and object category labels were established in this paper. These datasets were obtained by manually labeling based on existing datasets.
		The experimental results show that the proposed model achieves significantly improved performance for both seen and unseen testing objects.

		\item To maximize the collaborative learning of object category classification and grasp pattern recognition, this paper proposes a new loss function (JCEAR) from the Bayesian perspective. The regular parameters in the JCEAR can be learned adaptively during the training process. Based on JCEAR, a training strategy is formulated for training the parameters corresponding to the object category classification branch and the grasp pattern recognition branch in DcnnGrasp separately and iteratively. This makes the two branches teach each other interactively and improve the performance of both.

		\item 
  Experiments have been conducted to examine the robustness of the proposed method (DcnnGrasp) in obtaining 3D information of the objects on the dataset with missing grasp labels and its generalizability in unseen objects. All experimental results demonstrate the effectiveness of DcnnGrasp for all cases. Specifically, for the unseen problem, our method achieved a global accuracy of about 94\% and 99\% on the RGB-D Object dataset and Hit-GPRec dataset, respectively. Meanwhile, DcnnGrasp outperformed the second-best by nearly 15\% in a global accuracy on the RGB-D Object dataset, indicating that the proposed method can solve the unseen problem well. In addition,  when only one object with gesture
labels in each object category appears in the training process, the GA values of DcnnGrasp are still higher than 90\% and 95\% on the RGB-D Object dataset and Hit-GPRec
dataset  respectively, indicating the proposed method can solve the problem of difficult gesture labeling in grasp pattern recognition to some extent.

\end{itemize}
	\section{Related Work}\label{section:related work}
    In \cite{dovsen2011transradial}, Do{\v{s}}en \emph{et al.} integrated a simple vision system into the prosthetic hand, in which the camera (distance hardware) and software were used to recognize the object, and a control signal was generated for the prehension of the artificial hand. After that, a series of computer vision-based grasp pattern recognition methods have been proposed \cite{zandigohar2021netcut,wake2021object}.
For example, Kopicki \emph{et al.} \cite{kopicki2016one} proposed a one-shot learning method of dexterous grasps for the case of novel objects, in which point cloud image data were collected by a depth camera (RGB-D camera). In \cite{kopicki2016one}, the model of each grasp type was learned from a single kinesthetic demonstration. Then, multiple models learned from a single kinesthetic demonstration for each grasp type were used for grasp prediction and generation. This improved the robustness of the method for the incomplete data in both the training and testing stages.

The popularity of mobile phones generally makes image data more accessible. Ghazaei \emph{et al.} \cite{RN2} was the first to apply grey images data of household objects to the grasp pattern recognition problem, in which CNN (only consists of two convolutional layers and a downsampling layer) was used to predict grasp type from the image of an object. 
Furthermore, Deng \emph{et al.} \cite{deng2019attention} developed an attention-based visual analysis framework which was used to obtain grasp-relevant information in a scene. A deep convolutional neural network was used for detecting the grasp type and locating the attention point on the object. This method speeded up grasp planning with more stable configurations.
To improve the effectiveness of grasp pattern recognition, Shi \emph{et al.} \cite{shi2020computer} combined the 3D information and shape of the object using the depth image and grayscale image simultaneously.
Corona \emph{et al.} \cite{corona2020ganhand} proposed another approach to learn the 3D information of the objects in the image effectively. They designed a multi-task GAN architecture called GanHand to estimate the 3D shape of the objects from a single RGB image and predicts the best grasp type according to the taxonomy with 33 grasp types \cite{corona2020ganhand}.

Considering the case of embedding the camera in the palm of a robotic hand (or prosthetic hand), the identified grasp types for the same object from different approaching orientations of hand may be different \cite{han2019hand, zandigohar2019towards}. Therefore, in \cite{han2019hand}, instead of making absolute positive or negative predictions, Han \emph{et al.} utilized the priority of grasp type as the form of prediction result, and took the ranked lists of grasps as a new label form. Based on this, they constructed a probabilistic classifier using the Plackett-Luce model. Furthermore, Zandigohar \emph{et al.} \cite{zandigohar2019towards} used the probability lists of grasps as the form of prediction result.
However, this method does not utilize category information, so it requires more training data to cover the feature space to achieve high performance.
	
	\section{The Proposed Learning Methods}\label{section:proposedd method}

	In this part, the design of our proposed method (including DcnnGrasp and JCEAR) for joint learning of object category classification and grasp pattern recognition to achieve grasp pattern recognition performance. 
	
	\subsection{DcnnGrasp for Grasp Pattern Recognition}
	To achieve the goal of learning object category classification and grasp pattern recognition jointly, this paper designs a dual-branch network structure (DcnnGrasp).
		As shown in Figure ~\ref{new_overview-network}, DcnnGrasp consists of two branches, both of which take a given household object image $\textbf{x}$ as the input. For the convenience of presentation, this paper uses `category' and `grasp' to respectively represent the upper branch and the lower branch in Figure ~\ref{new_overview-network} according to their function. Each branch consists of one feature extractor and one classifier, and the features extracted from the object category feature extractor and the grasp feature extractor are integrated to enhance the performance of grasp pattern recognition. For grasp feature extractor, CnnGrasp is used in this paper. In the following, the object category feature extractor, the category classifier, and the grasp classifier are introduced in detail.
			\subsubsection{Object category feature extractor}
		The object category feature extractor consist of convolutional layers (DenseNet) and $m_{\mathrm{int}}$ fully connection layers. Assume that $f_{\mathrm{category}}(\textbf{x})$ is the feature maps obtained by convolutional layers, and define $\textbf{h}^{[0]}_{\mathrm{category}}\in \mathbb{R}^{d^{[0]}_{\mathrm{category}}}$ as vectorized $f_{\mathrm{category}}(\textbf{x})$ for convenience.
		 Let
	$$\textbf{h}^{[m]}_{\mathrm{category}}=g(\textbf{W}_{\mathrm{category}}^{[m]}\textbf{h}^{[m-1]}_{\mathrm{category}}+\textbf{b}^{[m]}_{\mathrm{category}})\in \mathbb{R}^{d^{[m]}_{\mathrm{category}}}$$
	be the output of the $m$-th layer with a dimension of $d^{[m]}_{\mathrm{category}}$ ($m=1,2,..,m_{\mathrm{int}}$), where $g(\cdot)$ is a nonlinear activation function.

		Therefore, the object category features $\textbf{I}_{\mathrm{category}}$ can be obtained by
$$
	    \textbf{I}_{\mathrm{category}} = g(\textbf{W}_{\mathrm{category}}^{[m_{\mathrm{int}}]}\textbf{h}^{[m_{\mathrm{int}}-1]}+\textbf{b}^{[m_{\mathrm{int}}]})\in \mathbb{R}^{d^{[m_{\mathrm{int}}]}_{\mathrm{category}}},
$$
Here,  $\theta_1$ represents the learnable parameters in $f_{\mathrm{category}}(\textbf{x})$ and the parameters $$\left\lbrace\textbf{W}_{\mathrm{category}}^{[m]},\textbf{b}_{\mathrm{category}}^{[m]}\right\rbrace_{m=1}^{m_{\mathrm{int}}}.$$

		\subsubsection{Category classifier}
	The category classifier takes $\textbf{I}_{\mathrm{category}}$ as input. Let
	$$\textbf{h}^{[m]}_{\mathrm{category}}=g(\textbf{W}_{\mathrm{category}}^{[m]}\textbf{h}^{[m-1]}_{\mathrm{category}}+\textbf{b}^{[m]}_{\mathrm{category}})\in \mathbb{R}^{d^{[m]}_{\mathrm{category}}}$$
	be the output of the $(m-m_{\mathrm{int}})$-th layer in the category classifier  ($m=1+m_{\mathrm{int}},2+m_{\mathrm{int}},..,m_{\mathrm{category}}$). The output of the category classifier can be obtained as:
    	   $$ \textbf{f}^o(\textbf{x};\theta_1,\theta_2) =g(\textbf{W}_{\mathrm{category}}^{[m_{\mathrm{category}}]}\textbf{h}^{[m_{\mathrm{category}}-1]}_{\mathrm{category}}+\textbf{b}^{[m_{\mathrm{category}}]}_{\mathrm{category}}),$$
  where $\theta_2$ stands for the parameters $$\left\lbrace\textbf{W}_{\mathrm{category}}^{[m]},\textbf{b}_{\mathrm{category}}^{[m]}\right\rbrace_{m=1+m_{\mathrm{int}}}^{m_{\mathrm{category}}}.$$
	
		\subsubsection{Grasp classifier}
	
	Assuming that $\textbf{I}_{\mathrm{grasp}}$ is the output of the grasp type feature extractor. The grasping information and object category information can be fused by concatenating $\textbf{I}_{\mathrm{grasp}}$ and $\textbf{I}_{\mathrm{category}}$, as shown in Fig.~\ref{new_overview-network}, and define the resulting feature as $\textbf{I}$.
	
	The grasp classifier takes $\textbf{I}$ as its input. This paper defines $\textbf{h}^{[0]}_{\mathrm{grasp}}\in \mathbb{R}^{d^{[0]}_{\mathrm{grasp}}}$ as  $\textbf{I}$
	for convenience.  Let
	$$\textbf{h}^{[m]}_{\mathrm{grasp}}=g(\textbf{W}_{\mathrm{grasp}}^{[m]}\textbf{h}^{[m-1]}_{\mathrm{grasp}}+\textbf{b}^{[m]}_{\mathrm{grasp}})\in \mathbb{R}^{d^{[m]}_{\mathrm{grasp}}}$$
	be the output of the $m$-th layer with a dimension of $d^{[m]}_{\mathrm{grasp}}$ ($m=1,2,..,m_{\mathrm{grasp}}$), where $g(\cdot)$ is a nonlinear activation function. The output of the category classifier can be obtained as:
    	   $$ \textbf{f}^{\mathrm{g}}(x;\theta_1,\theta_3,\theta_4) =g(\textbf{W}_{\mathrm{grasp}}^{[m_{\mathrm{grasp}}]}\textbf{h}^{[m_{\mathrm{grasp}}-1]}_{\mathrm{grasp}}+\textbf{b}^{[m_{\mathrm{grasp}}]}_{\mathrm{grasp}}),$$
  where $\theta_3$ and  $\theta_4$ represent the learnable parameters in the grasp type feature extractor and the parameters $$\left\lbrace\textbf{W}_{\mathrm{grasp}}^{[m]},\textbf{b}_{\mathrm{grasp}}^{[m]}\right\rbrace_{m=1}^{m_{\mathrm{grasp}}},$$
    	respectively.


	\subsection{Joint Cross-Entropy with Adaptive Regularizer}

In this paper, a novel loss function is developed to help the model to further extract and utilize the relationship between object category classification and grasp pattern recognition, thus improving the performance of the model on grasp pattern recognition.

 Let one-shot vectors $\textbf{c}^{\mathrm{g}}$  and $\textbf{c}^{\mathrm{o}}$ be the true labels for grasp pattern recognition and object category classification, respectively. $f_i^{\mathrm{g}}(\textbf{x};\theta_1,\theta_3,\theta_4)$ and $f_j^{\mathrm{o}}(\textbf{x};\theta_1,\theta_2)$  are the resulting probability of $x$ belonging to grasp type $i$ and object class $j$ respectively. Thus, $f_i^{\mathrm{g}}(\textbf{x};\theta_1,\theta_3,\theta_4)=p(c_i^{\mathrm{g}}=1|\textbf{x},\theta_1,\theta_3,\theta_4)$ and $f_j^{\mathrm{o}}(\textbf{x};\theta_1,\theta_2)=p(c_j^{\mathrm{o}}=1|\textbf{x},\theta_1,\theta_2)$.
	
	 To jointly lean multiple tasks,
	one simplest choice for the loss function is to use
	the following joint cross-entropy (JCE):
	\begin{align}\label{JCE}
    l_{\mathrm{JCE}}(\theta_k,\textbf{x}) =  -\sum_{i=1}^{l_\mathrm{o}}c_i^{\mathrm{o}}&\ln (f_i^{\mathrm{o}}(\textbf{x};\theta_1,\theta_2))\notag \\
    & - \sum_{j=1}^{l_\mathrm{g}}c_j^{\mathrm{g}}\ln (f_j^{\mathrm{g}}(\textbf{x};\theta_1,\theta_3,\theta_4)).
	\end{align}
	However, JCE is just the weighted sum of the cross-entropy functions of the two tasks, which loses the relationship between object category classification and grasp pattern recognition. To achieve the goal of jointly learning object category classification and grasp pattern recognition, this paper proposes Joint Cross-Entropy with Adaptive Regularizer (JCEAR) from the Bayesian
perspective.

	Since each of $\textbf{c}^{\mathrm{g}}$ and $\textbf{c}^{\mathrm{o}}$ follows a multinomial distribution, the following likelihood function can be taken:
	\begin{align}\label{MAP}
	p(\textbf{c}^{\mathrm{g}},\textbf{c}^{\mathrm{o}}|\theta_k,\textbf{x})&=p(\textbf{c}^{\mathrm{g}}|\theta_1,\theta_3,\theta_4,\textbf{x})p(\textbf{c}^{\mathrm{o}}|\theta_1,\theta_2,\textbf{x})\notag\\
	&=\prod_{i=1}^{l_\mathrm{o}}f_i^{\mathrm{o}}(\textbf{x};\theta_1,\theta_2)^{c_i^{\mathrm{o}}}\prod_{j=1}^{l_\mathrm{g}}f_j^{\mathrm{g}}(\textbf{x};\theta_1,\theta_3,\theta_4)^{c_j^{\mathrm{g}}}.
	\end{align}
	Let $ \textbf{W}\in \mathbb{R}^{l_\mathrm{o}\times l_\mathrm{g}}$ be a matrix, in which each element of $\textbf{W}$ is  $W_{i,j}=p(c^{\mathrm{o}}_i=1|c^{\mathrm{g}}_j=1,\textbf{x},\theta_k)(k=1,2,3,4)$. Then,
	\begin{align}
&\sum_{j=1}^{l_\mathrm{g}}  W_{i,j}f_j^{\mathrm{g}}(\textbf{x};\theta_1,\theta_3,\theta_4)  \notag\\
=&\sum_{j=1}^{l_\mathrm{g}} p(c^{\mathrm{o}}_i=1, c^{\mathrm{g}}_j=1,\textbf{x},\theta_k)p(c^{\mathrm{g}}_j=1|\textbf{x},\theta_1,\theta_3,\theta_4)\notag\\
	=&\sum_{j=1}^{l_\mathrm{g}} p(c^{\mathrm{o}}_i=1,c^{\mathrm{g}}_j=1|\textbf{x},\theta_k)
	=p(c^{\mathrm{o}}_i=1|\textbf{x},\theta_1,\theta_2)\notag\\
	&~~~~~~~~~~~~~~~~~~~~~~~~~~~~~~~~=f_i^{\mathrm{o}}(\textbf{x};\theta_1,\theta_2).
	\end{align}
	Therefore
	$ \textbf{W} \textbf{f}^{\mathrm{g}}(\textbf{x};\theta)= \textbf{f}^{\mathrm{o}}(\textbf{x};\theta).$

	By calculating the distribution of $\textbf{c}^o$ and $\textbf{c}^g$ over the training set, $W_{i,j}=p(c^{\mathrm{o}}_i=1|c^{\mathrm{g}}_j=1,\textbf{x},\theta_k)(k=1,2,3,4)$ can be estimated as
	\begin{align}\label{w}
	\hat{W}_{i,j} = \frac{N_{i,j}}{\sum_{q=1}^{l_g} N_{i,q}}
	\end{align}
	 where $N_{i,j}$ is the number of the images belonging to the $i$-th class object and the $j$-th class grasp.

	Assume that $\textbf{f}^{\mathrm{o}}(\textbf{x};\theta)-\hat{\textbf{W}}\textbf{f}^{\mathrm{g}}(\textbf{x};\theta)\sim \mathcal N(\textbf{0}, \bm{\sigma })$, where $\bm{\sigma   }=[\sigma_1, \sigma_2,...,\sigma_{l_o}]$.
	Therefore, define a prior
	\begin{align}  p(\theta_k,\textbf{x})=\frac{1}{\sqrt{2\pi }\sigma _1...\sigma_{l_o}}\exp^{\sum_{i=1}^{l_o} -\frac{\Gamma_i(\textbf{x})^2}{2\sigma_i^2}},
	\end{align}
	where
	    \begin{equation}
    \label{sigma-t 1}
      \Gamma_i(\textbf{x}) = \textbf{f}_i^o(\textbf{x};\theta_1;\theta_3;\theta_4) - \hat{\textbf{W}}_{i,:}\textbf{f}^g(\textbf{x};\theta_1;\theta_2),
    \end{equation}
	and $\hat{\textbf{W}}_{i,:}$ is $i$-th row vector of $\hat{\textbf{W}}$. Then, from the Bayesian perspective, $\hat {\theta}_k$ can be obtained by
	\begin{align}\label{MAP2}
	&\arg\max_{\theta_k} \ln (	p(\theta_k,\textbf{x}|\textbf{c}^{\mathrm{g}},\textbf{c}^{\mathrm{o}}))\notag\\
	=&\arg\max_{\theta_k}\ln (	p(\textbf{c}^{\mathrm{g}},\textbf{c}^{\mathrm{o}}|\theta_k,\textbf{x})p(\theta_k,\textbf{x}))\notag\\
	=&\arg\min_{\theta_k} l_{\mathrm{JCE}}(\theta_k,\textbf{x})+\frac{1}{2\sigma_1...\sigma _{l_o}}\sum_{i=1}^{l_o}\Gamma_i(\textbf{x})^2.
	\end{align}
	This leads to the minimization objective $l(\theta_k)$ as the loss function for the proposed model, where
	\begin{align}\label{JCEAR-V}
	l(\theta_k)&=l_{\mathrm{JCE}}(\theta_k,\textbf{x})+\frac{1}{2\sigma_1...\sigma_{l_o}}\sum_{i=1}^{l_o}\Gamma_i(\textbf{x})^2.
	\end{align}
 Meanwhile, the parameter $\sigma$ in the proposed loss function JCEAR is updated adaptively during the training process according to the following rule: \\
	$\bm{\sigma }$ is initialized as $\bm{\sigma }^{(0)}=[+\infty,\cdots,+\infty]$, and $\bm{\sigma }$ is updated by \begin{align}\label{updateSigma}
	\bm{\sigma }^{(t)} = \frac{1}{K}\sum_{k=1}^K \bm{\sigma }^{(t)}_k,
	\end{align}
		where $t$ is the number of iterations (See Algorithm \ref{BCD}), $\bm{\sigma }^{(t)}_k = [\sigma^{(t)}_{k,1},\sigma^{(t)}_{k,2}...,\sigma ^{(t)}_{k,l_0}] $, and
			\begin{equation}
	   \label{sigma-t_3}
	   \sigma^{(t)}_{k,i} = \sqrt{\frac{1}{H}\sum_{h=1}^{H}(\Gamma_i(\textbf{x}_h) - \alpha_i)^{2}},
	\end{equation}
	and $\textbf{x}_h$ is $h$-th image in the $k$-th sample batch.
Here,
    \begin{equation}
	   \label{sigma-t_2}
	   \alpha_i = \frac{1}{H}\sum_{h=1}^{H} \Gamma_i(\textbf{x}_h).
	\end{equation}
	
\begin{algorithm}[t]
		\caption{Training strategy based on JCRAR}\label{BCD}
		\textbf{Initialize: } $\bm{\sigma }^{(0)}=[+\infty,\cdots,+\infty]$, $t=0$, the learnable parameters in $f_{\mathrm{category}}(\textbf{x})$ by pretraining DenseNet with ImageNet \cite{deng2009imagenet}, the remaining parameters $\left\lbrace\textbf{W}_{\mathrm{category}}^{[m]},\textbf{b}_{\mathrm{category}}^{[m]}\right\rbrace_{m=1}^{m_{\mathrm{int}}}$ within $\theta_1^{(0)}$  and $\theta_2^{(0)}$ randomly.  \\
	
		Computing $\hat W$ by \eqref{w};\\
		\textbf{While} not converged \textbf{do} \\
			1. Fix $\theta_3^{(t)}$ and $\theta_4^{(t)}$, and update $\theta_1^{(t+1)}$ and $\theta_2^{(t+1)}$ by
			\begin{equation}\label{step1}	  \arg\min_{\theta_3,\theta_4}l(\theta_1^{(t)},\theta_2^{(t)},\theta_3,\theta_4)
		\end{equation}\\
		2. Update $\bm{\sigma }^{(t)}$ by \eqref{updateSigma};\\
		3. Fix $\theta_1^{(t+1)}$ and $\theta_2^{(t+1)}$, and update $\theta_3^{(t+1)}$ and $\theta_4^{(t+1)}$ by
		\begin{equation}\label{step2}
	 	\arg\min_{\theta_1,\theta_2}l(\theta_1,\theta_2,\theta_3^{(t+1)},\theta_4^{(t+1)})
		\end{equation} \\
		4. $t=t+1$;\\
		5. Check the convergence conditions\\
		$$| l(\theta_k^{(t+1)})-  l(\theta_k^{(t)})|\leq \epsilon.$$
		\textbf{end while}
		
	\end{algorithm}	

	The whole training learning strategy based on JCRAR is presented in
  Algorithm \ref{BCD}. To guarantee that each module in DcnnGrasp focuses on its own task, in
  Algorithm \ref{BCD}, the learnable parameters of the two branches are updated  separately\footnote{Different from deep mutual learning \cite{zhang2018deep}, and the two branches are trained for their own task by a common loss function.}.



    \section{Experiments}\label{section:experiment}
		\subsection{Experimental Settings}
	\subsubsection{Dataset}
	To evaluate the effectiveness of the methods, five datasets including the RGB-D Object dataset 
	\cite{lai2011large}, Hit-GPRec dataset 
	\cite{shi2020computer}, Amsterdam library of object images \cite{geusebroek2005amsterdam}, Columbia Object Image Library (COIL-100) \cite{nene1996columbia}, and MeganePro dataset 1 were used in the experiment.

\begin{table*}[ht]
\centering
\caption{Object category-based sampling method (OCS).}\label{dataset divided}
\begin{tabular}{ccc|cc|c}
\hline
\multirow{2}{*}{\begin{tabular}[c]{@{}c@{}}Total objects number\\ in $j$-th object category\end{tabular}}    &                                                  &           \multirow{2}{*}{\begin{tabular}[c]{@{}c@{}}The
expected number of \\ objects with grasp labels\end{tabular}}         &  \multicolumn{2}{c|}{Training Set + Validation Set} & \multirow{2}{*}{Testing Set}  \\ \cline{4-5}
                   &   &                 & With grasp labels & Without grasp labels &   \\ \hline
\multirow{2}{*}{$m_j$}&   $\leq$ &$ n_j$   & $m_j$     &         $0 $             & $0 $       \\
                                          &                   $>$                &  $n_j$ & $n_j $ &$m_j-n_j-1$                & $1$        \\ \hline
\end{tabular}
\end{table*}
	
	\begin{itemize}	
		\item \textbf{RGB-D Object dataset} \footnote{\url{http://rgbd-dataset.cs.washington.edu/dataset.html}} contains 300 objects (aka instances) in 51 categories with a total of 207921 images.
	The dataset was labeled \footnote{\url{https://drive.google.com/file/d/1PqQ_5HJOmUcQHLNnBUPUD6_qqhR7s-3c/view}} according to four grasp types widely used in our daily life including palmar wrist neutral, palmar wrist pronated, pinch, and tripod. A detailed description is presented in the supplementary materials of this paper.
		
		\item \textbf{Hit-GPRec dataset}\footnote{\url{http://homepage.hit.edu.cn/ydp
}}   contains 121 daily objects taken from different environmental conditions (4 types of lighting, 16 rotation angles of view, 4 different camera positions, and different postures of the objects). These objects can be categorized into four grasp types (including cylindrical, lateral, spherical, and tripod). Meanwhile, these objects were manually divided into 66 object categories further\footnote{\url{https://drive.google.com/file/d/14bEIDjd9a-80LKk0I2xnuQmUUhcNzLoi/view
}}.
    \item \textbf{Amsterdam library of object images (ALOI)}\footnote{\url{https://aloi.science.uva.nl/
}} contains colorful images of a thousand small objects. Each object has 72 images with a black background. This dataset was initially used for object category classification tasks. Then, Ghazaei \emph{et al.} \cite{RN2} chose 494 objects from the dataset and added grasp types labels (including four gestures: palmar wrist neutral, palmar wrist pronated, pinch, and tripod) to these objects\footnote{\url{https://iopscience.iop.org/article/10.1088/1741-2552/aa6802/data }}. Based on this, our study  manually divided the objects into 129 categories\footnote{\url{https://docs.google.com/spreadsheets/d/1qvuiRL0x8NTgjMNrRRpDo9utKgCeNFoH/edit\#gid=72548980}}.

    \item \textbf{Columbia University Image Library
(COIL-100)}\footnote{\url{https://www.cs.columbia.edu/CAVE/software/softlib/coil-100.php}} contains one hundred objects. Each object was photographed according to a 5-degree rotation, and 72 color images with a size of 128$\times$128 were obtained. Our study chose 96 objects that can be grasped with a single hand and manually labeled these images with four gestures (including palmar wrist neutral, palmar wrist pronated, pinch, and tripod) and 53 object category labels \footnote{\url{https://drive.google.com/file/d/1psxIYfG9t8Y-_WwO8lIiL4Ykg848qi2a/view?usp=sharing}}.

    \item \textbf{MeganePro dataset 1}\footnote{\url{https://dataverse.harvard.edu/dataset.xhtml?persistentId=doi:10.7910/DVN/1Z3IOM}} were obtained from 30 healthy subjects with an average age of 46.63 ± 15.11 years\cite{cognolato2020gaze}. It contains 18 object categories and 10 grasp types that were selected based on the hand taxonomies \cite{cutkosky1989grasp} and grasp frequency in Activities of Daily Living \cite{bullock2013grasp}. Based on this, the object images were obtained by detecting and cropping from the original video dataset\footnote{\url{https://github.com/MountStonne/CropedObjectsByFrames}}.
  	
    	\end{itemize}

\subsubsection{Cross-validation}
	To verify the generalizability and robustness of grasp pattern recognition, two forms of dataset division were considered: within-whole dataset cross-validation (WWC) and between-object cross-validation (BOC).
	For WWC, the whole dataset was sampled and divided into a training set, a validation set, and a testing set at the ratio of $8:1:1$. The dataset division method BOC was used to verify the ability of the models for unseen objects (the objects do not appear in the training set and validation set). In this case, an object and its views were either wholly seen or unseen.
	$90\%$ of the objects in the dataset were selected and divided into a training set and a validation set at the ratio of $9:1$. The remaining $10\%$ objects in the dataset were taken as the testing set.
	For fair comparisons, this study used standard ten-fold cross-validation, and the average results were reported.


\subsubsection{Object Category-based Sampling}
To better investigate the relationship between object categories and grasp labels, we proposed an object category-based sampling (OCS) method, ensuring that at least one object with grasp labels in each object category appeared in the training process. Mark the total number of the $j$-th object category in the dataset and the expected number of randomly chosen objects with grasp labels in the training process as $m_j$ and $n_j$, respectively. As shown in Table \ref{dataset divided}, if OCS is chosen, for each object category,  $n_j$ objects with grasp labels, $m_j-n_j-1$ objects without grasp labels will be used in the training process, and the remaining one will be taken as the testing data, when $n_j>m_j$. When $n\leq m$, all objects in that category and their grasp labels will appear in the training process. All samples used in the training process
were divided into training and validation sets according to 9:1, whether grasp labels were used or not. It was worth mentioning that, for OCS, the object category labels of all samples in both the training and validation sets were used, and the one-hot vectors composed of zeros were used as the `label' of samples with missing grasp labels during the training process. OCS can be used to test the prediction performance of the model for grasp labels of different objects in the same object category and investigate the relationship between object category and grasp labels better.  Each experiment on OCS was performed three times in which the samples for training, validation and testing were chosen randomly according to Table \ref{dataset divided}, and the average results were reported.

\subsubsection{Implementation Details}

	As shown in Fig.~\ref{new_overview-network}, RGB images of household objects were taken as the input of our network. The input images were resized into 224$\times$224 pixels. The input images of the category branch (each pixel is divided by 255) the grasp branch were normalized according to
	\begin{equation}
	\label{normalised_1}
	\textbf{x}_{\mathrm{normalised}}^r = \frac{\textbf{x}^r - \varepsilon^r}{\eta^r},
	\end{equation}
	where $\textbf{x}^r$ represents the $r$-th channel of the image $\textbf{x}$ (with size of $M\times N$),
	\begin{equation}
	\label{normalised_2}
	\varepsilon^r = \frac{1}{N+M}\sum_{n=1}^{N}\sum_{m=1}^{M}\textbf{x}^r_{n,m},
	\end{equation}
	and
	\begin{equation}
	\label{normalised_3}
	\eta^r = \sqrt{\frac{1}{N+M}\sum_{n=1}^{N}\sum_{m=1}^{M}(\textbf{x}^r_{n,m}-\varepsilon^r)}.
	\end{equation}

 \begin{table}[h]
 \caption{Description of state-of-art methods}
\label{description_method}
\begin{tabular}{l|l}
\hline
Model Name & Description \\ \hline
CnnGrasp & \begin{tabular}[c]{@{}l@{}}consisted of two convolution layers and\\ a downsampling layer\end{tabular} \\ \hline
\begin{tabular}[c]{@{}l@{}}EfficientNet\\ EfficientNet\_v2\end{tabular} & \begin{tabular}[c]{@{}l@{}}considered network depth, width, and resolution\\ to balance accuracy and speed\end{tabular} \\ \hline
Lightlayers & \begin{tabular}[c]{@{}l@{}}resorted Matrix Factorization to reduce\\ the complexity of the DNN models without\\ much loss in the accuracy\end{tabular} \\ \hline
GhostNet & \begin{tabular}[c]{@{}l@{}}used the Ghost module to upgrade existing\\ convolutional neural networks\end{tabular} \\ \hline
RegNet & \begin{tabular}[c]{@{}l@{}}was a low-dimensional design space composed\\ of simple and regular networks\end{tabular} \\ \hline
\end{tabular}
\end{table}

  \begin{table*}[ht]
 \centering
 \caption{Comparison with other benchmark methods under WWC sampling. }\label{com_method}
\begin{tabular}{c|ccc|ccc|ccc|ccc|ccc}
\hline
\multirow{2}{*}{Methods} & \multicolumn{3}{c|}{RGB-D Object dataset}        & \multicolumn{3}{c|}{Hit-GPRec dataset}              & \multicolumn{3}{c|}{ALOI} & \multicolumn{3}{c|}{COIL-100}                       & \multicolumn{3}{c}{MeganePro dataset 1} \\ \cline{2-16}
                         & GA             & MF1            & MCR             & GA              & MF1             & MCR              & GA      & MF1    & MCR     & GA              & MF1             & MCR              & GA          & MF1        & MCR         \\ \hline
CnnGrasp                 & 75.81          & 70.13          & 73.77          & 78.56           & 68.42           & 70.11           & 95.45   & 95.65  & 95.67  & 97.45           & 97.18           & 97.07           &    98.57    &   97.84    &    97.81   \\
GhostNet                 & 97.16          & 97.23          & 97.58          & 65.31           & 62.99           & 67.28           &   68.40      &   64.38     &    68.74    &      28.13           &   10.97              &        25.00         &     14.09        &      2.46      &     10.00       \\
RepVGG                   & 98.76          & 98.85          & 98.92          & 97.43           & 97.37           & 97.48           &  96.66   &    96.80    &   96.78    & 95.78           & 95.61           & 95.89           &      96.66       &     95.66       &     96.07       \\
EfficientNet             & 98.20          & 98.17          & 98.13          & 79.74           & 77.75           & 78.18           & 88.69   & 89.06  & 89.48  & 39.76           & 29.09           & 37.89           &   63.93     &    60.21   &    62.22   \\
EfficientNet\_v2         & 99.73          & 99.74          & 99.73          & 86.25           & 84.53           & 84.32           & 97.86   & 97.87  & 97.89  & 93.09           & 92.55           & 92.85           &   97.57     &   96.76    &    96.93   \\
RegNetX600               & 93.22          & 93.48          & 94.01          & 47.74           & 38.11           & 45.73           & 83.56   & 82.76  & 83.73  & 81.16           & 80.05           & 82.06           &    78.77   &  75.19 & 75.52      \\
RegNetY600               & 97.06          & 97.21          & 97.12          & 52.56           & 47.17           & 53.25           & 91.23   & 91.57  & 91.16  & 81.85           & 79.87           & 80.57           &    86.72    &   82.41    &   81.42    \\
LightLayers              & 85.88          & 85.61          & 85.70          & 78.33           & 76.70           & 79.01           & 55.79   & 49.20  & 54.80  & 73.36           & 71.01           & 73.68           &   86.72     &     82.41  &    81.42  \\
DcnnGrasp                & \textbf{99.99} & \textbf{99.99} & \textbf{99.99} & \textbf{100.00} & \textbf{100.00} & \textbf{100.00} &   \textbf{99.24}      &  \textbf{99.21}      &    \textbf{99.26}    & \textbf{100.00} & \textbf{100.00} & \textbf{100.00} &   \textbf{98.66}     &   \textbf{98.53}   &    \textbf{98.60}   \\ \hline
\end{tabular}
\end{table*}

    \begin{table*}[h]
 \centering
  \caption{Comparison with other benchmark methods under BOC sampling. }\label{boc_method}
\begin{tabular}{c|ccc|ccc|ccc}
\hline
\multirow{2}{*}{Methods}          & \multicolumn{3}{c|}{RGB-D Object dataset}        & \multicolumn{3}{c|}{Hit-GPRec dataset}            & \multicolumn{3}{c}{ALOI}                         \\ \cline{2-10}
                 & GA             & MF1            & MCR             & GA             & MF1            & MCR             & GA             & MF1            & MCR             \\ \hline
CnnGrasp         & 70.70          & 70.88          & 71.16          & 81.59         & 80.61          & 81.67          & 71.77          & 72.17          & 72.06          \\
GhostNet         & 75.52          & 75.43          & 76.21          & 60.66          & 57.43          & 64.62          &       58.08         &       53.66        &       60.25         \\
RepVGG           & 72.60          & 72.67          & 73.90          & 93.87          & 93.68          & 94.01          &      74.08          &        74.96        &       75.32         \\
EfficientNet     & 79.07          & 78.67          & 80.07          & 73.56          & 69.75          & 70.13          & 72.38          & 72.42          & 74.17          \\
EfficientNet\_v2 & 76.84          & 75.65          & 76.59          & 78.27          & 76.46          & 75.19          & 72.49          & 73.07          & 75.66          \\
RegNetX600       & 69.61          & 67.93          & 69.59          & 55.99          & 51.54          & 54.62          & 68.71          & 68.69          & 69.87          \\
RegNetY600       & 69.30          & 70.17          & 71.80          & 62.40          & 57.22          & 57.96          & 70.20          & 70.22          & 72.13          \\
LightLayers      & 64.02          & 63.99          & 67.32          & 66.12          & 61.32          & 65.18          & 40.19          & 33.73          & 42.61          \\
DcnnGrasp        & \textbf{94.43} & \textbf{94.40} & \textbf{95.28} & \textbf{99.81} & \textbf{99.81} & \textbf{99.88} & \textbf{78.99} & \textbf{79.47} & \textbf{80.18} \\ \hline
\end{tabular}
\end{table*}

	The network structure is follows. The object category feature extractor adopts DenseNet121 \cite{huang2017densely} and the category classifier adopts three fully connected layers and a softmax layer. The final output dimensions of each fully connected layer are 256, 128, and 64, respectively. The softmax layer is essentially a fully connected layer, but the activation function adopts softmax. The final output dimension equals the number of object categories. The grasp type feature extractor adopts CnnGrasp. The grasp classifier is composed of two fully connected layers and a softmax layer. The final output dimension of each fully connected layer is 128 and 64. The final output dimension of the softmax layer equals the number of grasp types.
	
 For the optimization of the sub-problems \eqref{step1} and \eqref{step2}, Adam was used as the optimizer, and its learning rate was initialized as 0.001. The exponential decay rate for the 1st and 2nd moment estimates were set as 0.9 and 0.999, respectively.

 \subsection{Comparison with State-of-the-Art Methods}

\subsubsection{Evaluation Metrics and Compared Methods}
 	This study used the global accuracy (GA), the macro recall (MRC) \cite{thabtah2009naive} and the macro F1 (MF1) \cite{opitz2019macro} to evaluate the performance of the proposed method \cite{robert2014machine}.

 In this paper, our method was compared with the following state-of-art methods: CnnGrasp \cite{RN2}, EfficientNet  \cite{tan2019efficientnet}, EfficientNet\_v2 \cite{tan2021efficientnetv2}, Lightlayers  \cite{jha2020lightlayers}, GhostNet \cite{han2020ghostnet}, and RegNet \cite{radosavovic2020designing}. These methods are described in Table \ref{description_method}, where only CnnGrasp is designed specifically for grasp classification and the remaining five methods are given for the classical classification problem.

\subsubsection{Comparison in WWC}

To verify the robustness of the proposed method (DcnnGrasp) in different views of objects, our method was compared with all six methods in WWC on all five datasets. All experimental results are presented in Table \ref{com_method}. It can be seen from this table that DcnnGrasp achieved the best prediction performance compared with other methods for grasp classification.
Specifically, DcnnGrasp obtained an accuracy of almost 100\% on the RGB-D Object dataset, Hit-GPRec dataset, and COIL-100. For the Hit-GPRec dataset and COIL-100, the proposed method even outperformed most comparing methods by $15\%$ in accuracy.

The comparison of the grasp classification methods (including CnnGrasp and DcnnGrasp) with five classical classification methods indicated that both two grasp classification methods achieved excellent grasp classification performance on ALOI, COIL-100, and MeganePro datasets. For the RGB-D Object dataset and Hit-GPRec dataset, DcnnGrasp still achieved the best prediction performance, but the performance of CnnGrasp was mediocre. The difference between the accuracy of CnnGrasp and DcnnGrasp was even more than $24\%$ and $21\%$ on the RGB-D Object dataset and the Hit-GPRec dataset, respectively. All experimental results shown in Table \ref{com_method} illustrate the effectiveness of the proposed method in WWC.

\subsubsection{Comparison in BOC}

		\begin{figure*}[ht]
        \centering
        	\subfigure[GA on RGB-D Object dataset]{
		\includegraphics[scale=0.22]{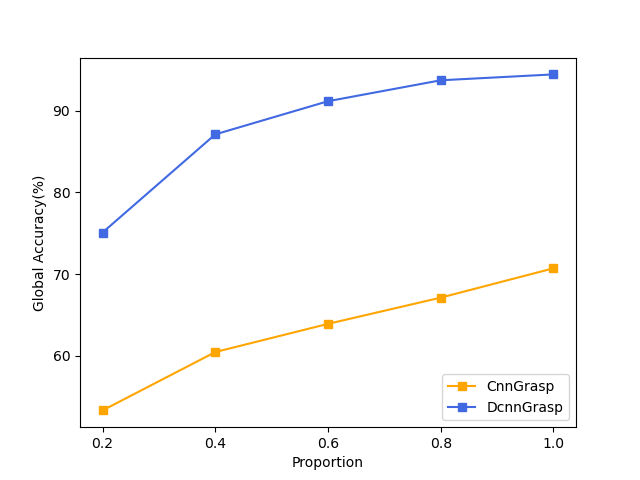}}
			\subfigure[GA on Hit-GPRec dataset]{
		\includegraphics[scale=0.22]{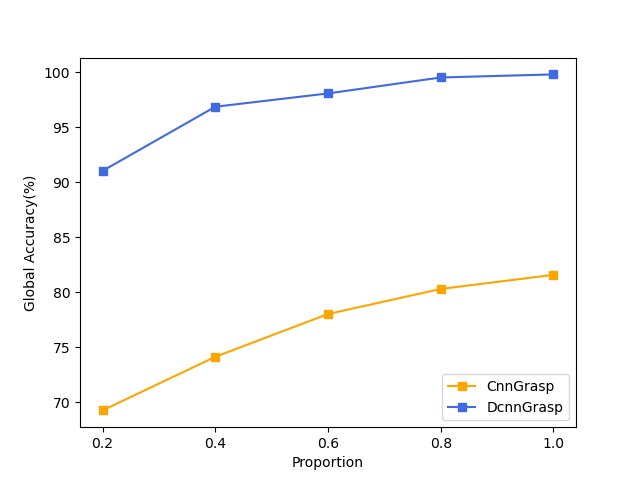}}
			\subfigure[GA on ALOI dataset]{
		\includegraphics[scale=0.22]{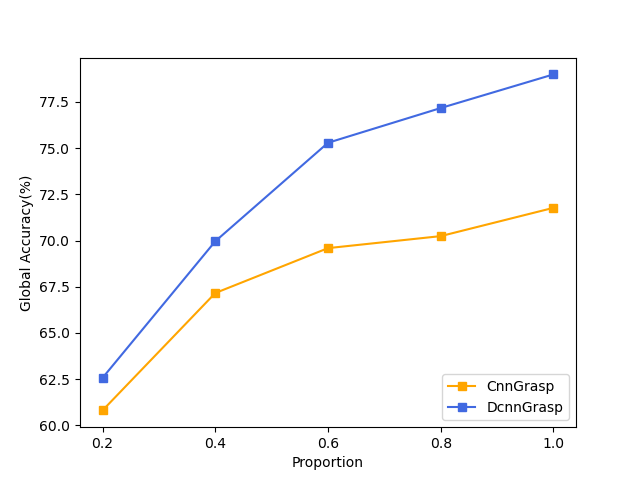}}
		
		
		
		\caption{GA 
		under different proportions (proportion means the ratio of data items with grasp labels to the data items in the entire dataset).}\label{weak}
		\end{figure*}

To demonstrate the generalizability of the proposed method in unseen objects, our method was compared with all six methods in BOC on three datasets including the RGB-D Object dataset, Hit-GPRec dataset, and ALOI (For BOC, only the datasets containing more than 100 objects were used). All results are presented in Table \ref{boc_method}. It can be seen from this table that DcnnGrasp still archived the best prediction performance for BOC.

In addition, the comparison of the results in Table \ref{com_method} and Table \ref{boc_method} indicated that BOC (unseen problem) was much more challenging than WWC. The performance of all compared methods decreased dramatically in most cases. On the RGB-D Object dataset, the gap between the results of WWC and BOC was even 20\% for some methods (such as GhostNet,  RepVGG, EfficientNet\_v2, RegNetX600, and RegNetY600). However, even for BOC, on the Hit-GPRec dataset, the performance of DcnnGrasp on all three evaluation metrics was close to 100\%. On the RGB-D Object dataset, they achieved an accuracy of more than 94\%, which is 15\% higher than that of all comparing methods.

\subsection{Robustness on the dataset with missing grasp labels}\label{sec_weak}

\subsubsection{Comparison in BOC}

  In this part, the robustness of DcnnGrasp was investigated on the dataset with missing grasp labels, in which the proportion of the remaining grasp labels was $p=\{0.2, 0.4, 0.6, 0.8, 1\}$. Meanwhile, one-hot vectors composed of zeros were used for losing grasp labels during the training process. DcnnGrasp was compared with the grasp classification method CnnGrasp on these datasets including the RGB-D Object dataset, Hit-GPRec dataset, and ALOI dataset for BOC sampling. All comparison results were illustrated in Fig. \ref{weak}.

   It can be seen from Fig. \ref{weak} that the performance of DcnnGrasp outperformed CnnGrasp for all cases significantly. In addition, we discovered that,  even for a small $p$, DcnnGrasp maintained an excellent performance on the RGB-D Object dataset and Hit-GPRec dataset. For $p=0.4$, the GA values of DcnnGrasp were still higher than 85\% and 95\% on the RGB-D Object dataset and Hit-GPRec dataset, respectively. It showed the robust grasping prediction ability of DcnnGrasp when datasets had more sufficient objects per object category, which guarantee DcnnGrasp can study the relationship between grasp types and object categories well.

\subsubsection{Comparison in OCS}
\begin{table}[]
\centering
    \caption{Experimental results under the proposed specific sampling method. Note that, there are $n$ objects in each object category with grasp labels in the training and validation sets.}\label{xianzhiweak}
\begin{tabular}{cccccccc}
\hline
\multicolumn{2}{c}{\multirow{2}{*}{Datasets}}           & \multicolumn{3}{c}{RGB-D Object dataset} & \multicolumn{3}{c}{Hit-GPRec dataset} \\
\multicolumn{2}{c}{}                                    & GA           & MF1         & MRE         & GA          & MF1        & MRE\\ \hline
\multirow{2}{*}{n=3} & \multicolumn{1}{c|}{DcnnGrasp} & \textbf{95.53}        & \textbf{95.75}       & \textbf{96.11}       & \textbf{98.40}       & \textbf{97.27}      & \textbf{96.98}      \\
                       & \multicolumn{1}{c|}{CnnGrasp}  & 68.13        & 66.95       & 66.71       & 78.76       & 75.44      & 74.34        \\ \hline
\multirow{2}{*}{n=2} & \multicolumn{1}{c|}{DcnnGrasp} & \textbf{91.46}        & \textbf{92.33}       & \textbf{91.64}       & \textbf{97.71}       & \textbf{96.94}      & \textbf{97.22}      \\
                       & \multicolumn{1}{c|}{CnnGrasp}  & 63.31        & 61.72       & 60.97       & 76.79       & 73.21      & 70.84      \\ \hline
\multirow{2}{*}{n=1} & \multicolumn{1}{c|}{DcnnGrasp} & \textbf{93.24}        & \textbf{93.99}       & \textbf{93.70}       & \textbf{96.98}       & \textbf{97.00}      & \textbf{96.72}   \\
                       & \multicolumn{1}{c|}{CnnGrasp}  & 58.75        & 56.37       & 55.43       & 68.20       & 59.96      & 57.49    \\ \hline
\end{tabular}
\end{table}

Since prosthetic or robotic hands will be generally initialized before leaving the factory in which a well-designed dataset can be used during the training process, we used OCS to simulate such scenarios in this experiment, where the expected number of objects with grasp labels in each object category $n$ was set as $n=\{1, 2, 3\}$. The results for the RGB-D Object dataset and Hit-GPRec dataset were presented in Table \ref{xianzhiweak}. These two datasets were chosen as they have more sufficient objects per object category compared with other datasets such as ALOI.

    \begin{table}[t]
	\renewcommand\arraystretch{1.5}
	\centering
    \caption{Ablation study of our methods. Note that \textcolor{red}{red} and \textcolor{blue}{blue} appear in some places. a and b represent the GA of the object category classification task and the grasp pattern recognition task, respectively.}\label{ablation record table}
    \begin{tabular}{c|cccc}
    \hline
       & v1 & v2 & v3 \\ \hline
    Dual branch network &    \checkmark & \checkmark & \checkmark \\
    Using object category label &     & \checkmark & \checkmark \\
    Training strategy based on JCEAR &     &  & \checkmark \\ \hline \hline
    RGB-D Object dataset   & \textcolor{blue}{72.87} &  \textcolor{red}{43.25}/\textcolor{blue}{76.38} & \textbf{\textcolor{red}{91.50}/\textcolor{blue}{94.43}} \\
    Hit-GPRec dataset  & \textcolor{blue}{82.96} & \textcolor{red}{39.10}/\textcolor{blue}{85.43} & \textbf{\textcolor{red}{98.31}/\textcolor{blue}{99.81}} \\
    ALOI dataset   & \textcolor{blue}{63.79} & \textcolor{red}{15.72}/\textcolor{blue}{69.92}  & \textbf{\textcolor{red}{41.27}/\textcolor{blue}{78.99}} \\\hline
    \end{tabular}
    \end{table}

From Table \ref{xianzhiweak}, we can see that the DcnnGrasp maintains much stable performance compared with CnnGrasp as $n$ decreases. Even for the case when only one object with grasp labels in each object category appears in the training process, i.e., $n=1$, the GA values of DcnnGrasp were still higher than 90\% and 95\% on the RGB-D Object
dataset and Hit-GPRec dataset, respectively, while the performance of  CnnGrasp drops hugely if $n$ is decreased to $1$. It demonstrated the strong robustness of DcnnGrasp for different objects in the same category and the case of the dataset with missing grasp labels.



 	\subsection{Ablation Studies}

	\subsubsection{Effects of Different GTFEs}

		\begin{figure*}[]
        \centering
        	\subfigure[GA on RGB-D Object dataset]{
		\includegraphics[scale=0.25]{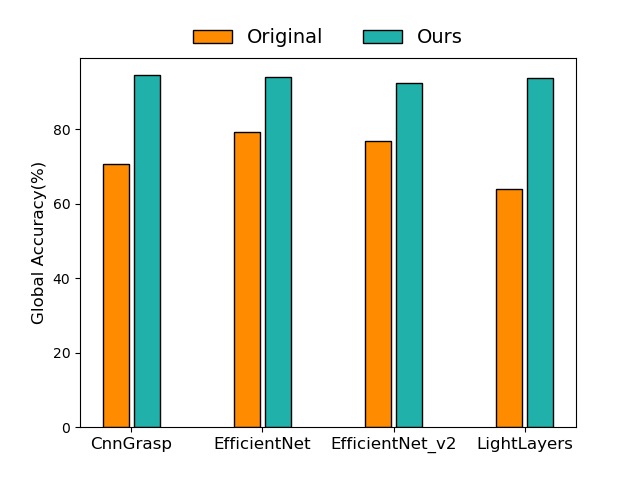}}
        	\subfigure[GA on Hit-GPRec dataset]{
		\includegraphics[scale=0.25]{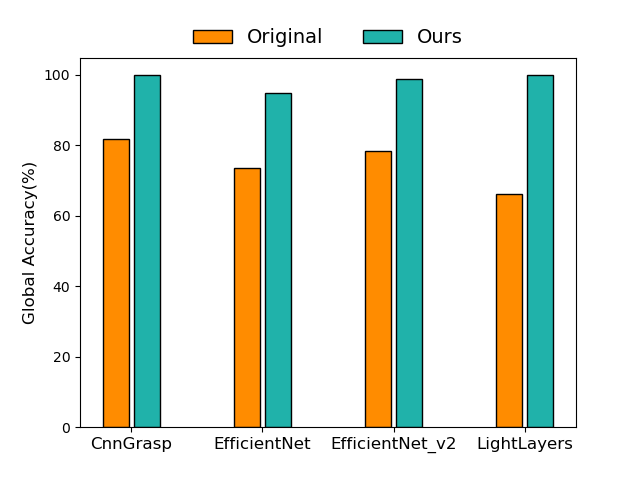}}
					\subfigure[GA on ALOI dataset]{
		\includegraphics[scale=0.25]{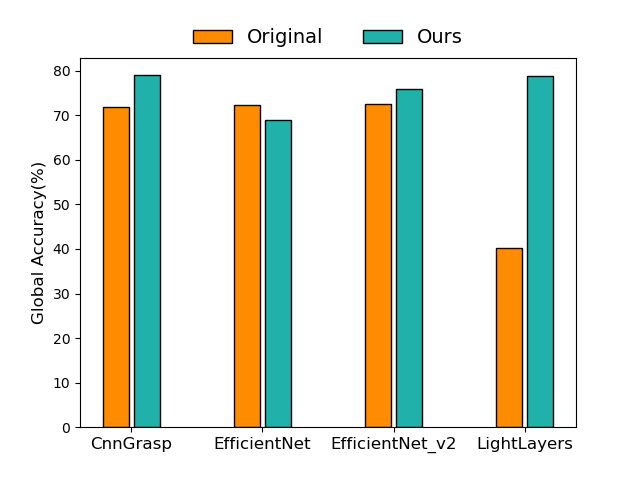}}
		\caption{Comparison on the effectiveness of four different grasp type feature extractors in the case of BOC sampling. `Original' represents four models, naemly CnnGrasp, EfficientNet, EfficientNet\_v2, and LightLayers. `Ours' represents DcnnGrasp using the four models as grasp type feature extractors. }\label{backbone}
	\end{figure*}
	
	Ablation studies were performed from two aspects: (1) grasp type feature extractor (GTFE), and (2) component of the proposed method. The impact of each aspect on the model performance was analyzed in turn.

	This study respectively used EfficientNet (B0), EfficientNet\_v2 (B0), LightLayers, and CnnGrasp as the grasp type feature extractor to investigate the impact of different extractors on the network performance\footnote{Three networks including EfficientNet (B0), EfficientNet\_v2 (B0), and LightLayers were chosen here because of their efficiency and good performance on comparison experiments.}. The experimental results are shown in Fig.~\ref{backbone}. It can be seen from this figure that the methods taking CnnGrasp as the grasp type feature extractor in DcnnGrasp achieved the best results, which shows that the choice of grasp type feature extractor is not related to the depth of the network (CnnGrasp only has two layers). Meanwhile, the performance of all four models including CnnGrasp, EfficientNet (B0), EfficientNet\_v2 (B0), and LightLayers was enhanced by the proposed method significantly in most cases. Specifically, for the RGB-D Object dataset and Hit-GPRec dataset, the GA values of all methods taking four traditional models as the grasp type feature extractor reached more than 90\%, which is about 20\% higher than those of the four models. As it will be discussed later, this performance advantage is not only benefited from the dual-branch network but also from the training strategy based on JCEAR, which enhanced the joint learning of object category classification and grasp pattern recognition significantly.


	\subsubsection{Effects of Different Strategies}

	To validate the effectiveness of the proposed strategies to improve the performance of grasp pattern recognition (including the dual-branch network and the introduction of object category information and training strategy based on JCEAR), this paper compared the following variants obtained by combining the strategies gradually:
	\begin{itemize}
  \item [v1:]
  The dual-branch network is used, and the cross-entropy is taken as the loss function.
  \item [v2.]
   The dual-branch network is used, and the joint cross-entropy is taken as the loss function, in which object category labels are used.
  \item [v3.] Both the dual-branch network and training strategy based on JCEAR are used, in which object category labels are used.

\end{itemize}
	All comparison results are presented in Table \ref{ablation record table}. It can be seen that the significant increase in GA is attributed to each proposed strategy, demonstrating the effectiveness of the proposed strategies. Meanwhile, the comparison of v1 and v2 indicates significant improvements attributed to the proposed training strategy based on JCEAR. Especially, on the RGB-D Object dataset and Hit-GPRec dataset, the increase in GA is at least 13.5\%.






	\begin{figure*}[t]
        \centering
        	\subfigure[The results on the case of WWC sampling.]{
		\includegraphics[scale=0.4]{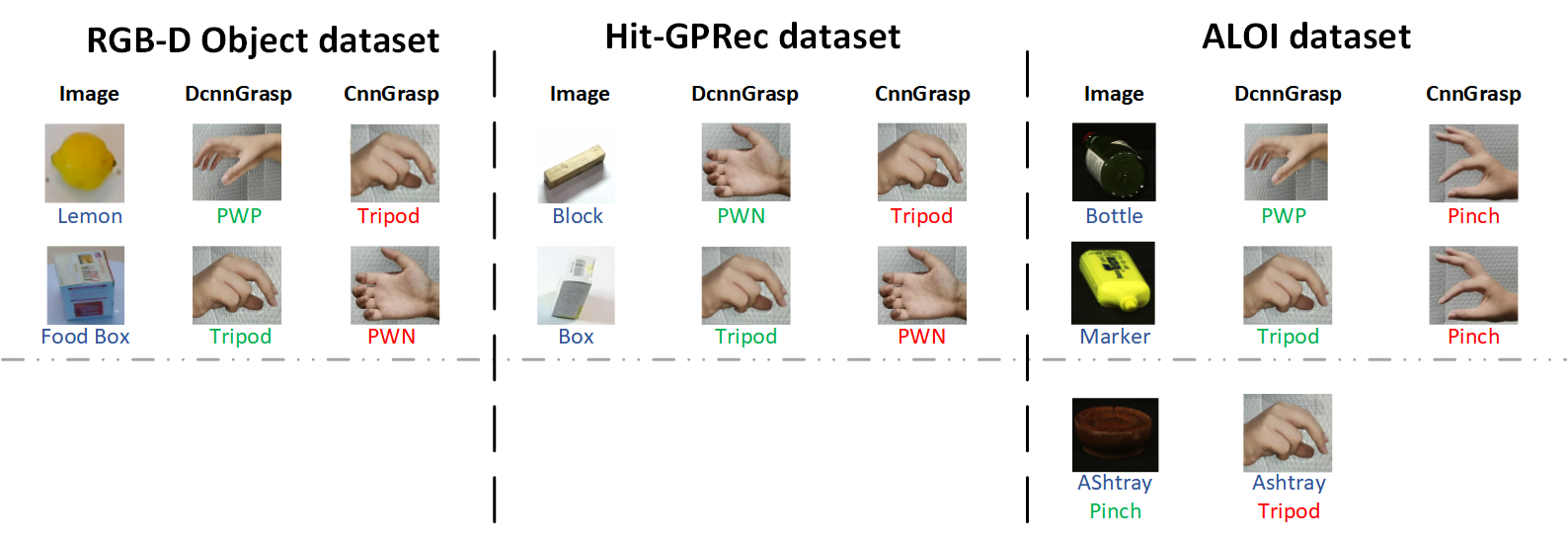}}
			\subfigure[The results on the case of BOC sampling.]{
		\includegraphics[scale=0.4]{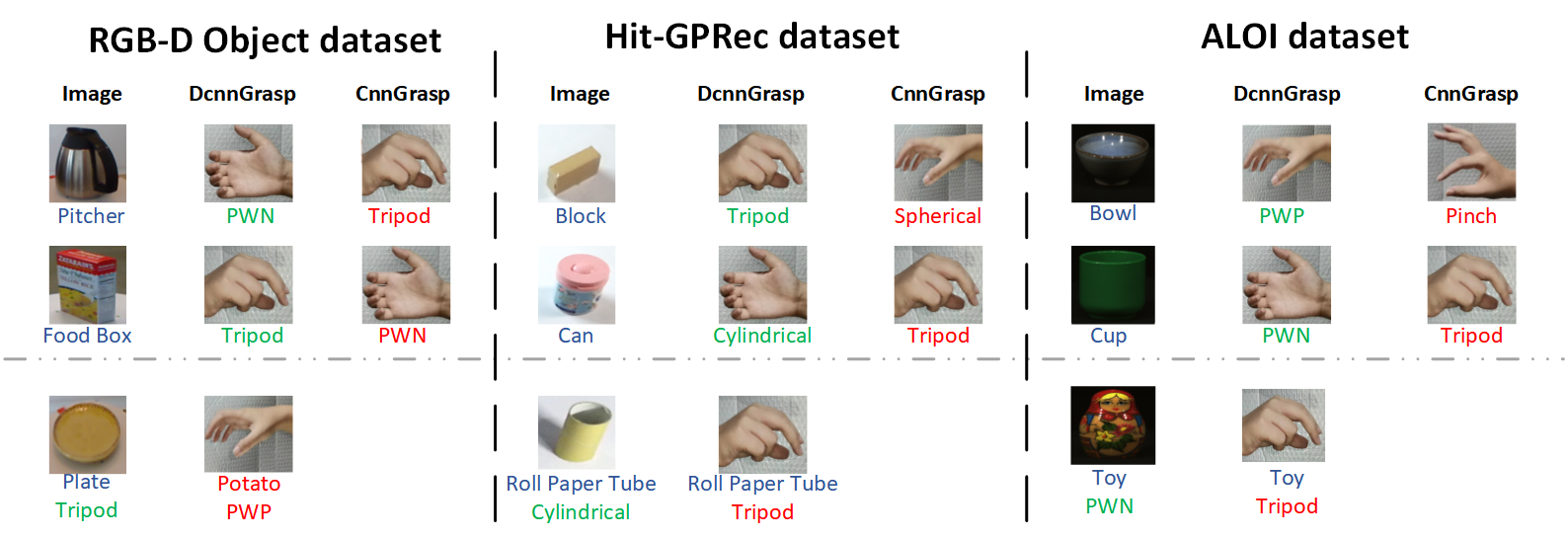}}
		\caption{Classified visual results. The blue label represents the object category label. The green label indicates a correct classification, and the red label indicates a wrong classification. }\label{vision}
	\end{figure*}

    \section{Discussion}\label{section:results and analysis}
    In the above experiments, for most cases, DcnnGrasp achieved the best performance.  To analyze the performance of DcnnGrasp in grasp pattern recognition deeply, based on the observations from the experiment and the visualized results of DcnnGrasp and CnnGrasp in Fig. \ref{vision}, this paper performs further discussions on the following four aspects: (1) robustness in 3D information of the objects for WWC, (2) sensitivity to the shadow and confusing background, (3) generalizability in unseen objects, and (4) robustness in a dataset with missing grasp labels.





    \subsection{Robustness in 3D information of the objects for WWC}

     From the comparison results in WWC, it can be seen that DcnnGrasp obtained the best results on all datasets, and all evaluation metric values were more than 98\%, indicating strong robustness of DcnnGrasp in different views of the objects. Meanwhile, from the visualized results shown in Fig. \ref{vision} (a), compared with DcnnGrasp, CnnGrasp obtained confusion about the 3d information of the objects easier. Specifically, in the prediction for `Lemon', CnnGrasp obtained a wrong grasping gesture `Tripod'. This is possibly because the 3d shape of `Lemon' was mistakenly recognized as flat. In addition, from the cases of `Food Box' and `Box', it can be seen that CnnGrasp was sensitive to the deformation generated by taking photos of the object from different angles. However, even for this situation, DcnnGrasp still worked well.
    All these observations show the strong robustness of DcnnGrasp in obtaining 3D information of the objects, which is attributed to the introduction of object category information.


    \subsection{Sensitivity to the shadow and confusing background}


   In grasp pattern recognition, shadows and confusing backgrounds often appear in household object image data, which will affect the performance of grasp classification. For example, in the ALOI dataset, the example images listed in Fig. \ref{vision} show that the target object is usually indistinguishable from the background. For `Bottle', `AShtray', and `Bowl', it is difficult to recognize the target object even for humans. This is one of the reasons why most methods failed on the ALOI dataset, particularly BOC sampling. Another example is the Hit-GPRec dataset contaminated by the shadows. As shown in Table \ref{com_method} and \ref{boc_method}, on this dataset, DcnnGrasp achieved a value of at least 99\% for the evaluation metrics in all cases. This good performance is attributed to the proposed training strategy based on JCEAR, as shown in the ablation study given in Table \ref{ablation record table}.

    \subsection{Generalizability in unseen objects}

   According to the comparison results given in Tables \ref{com_method} and \ref{boc_method}, most of the methods performed worse than WWC in BOC validation, indicating the considerable challenge of unseen problem. One of the reasons may be the dataset. For example, on the ALOI dataset, there are only a few different individuals in each category, and for some object categories, there is even one object. This makes it difficult to learn the relationship between object categories and grasping gestures. A similar situation can be observed on other datasets. For example, on the Hit-GPRec dataset, there are only two objects in the category `Roll Paper Tube', and the two objects are marked with different grasp types (one is `Tripod', and the other is `Cylindrical'), which leads to a wrong prediction by DcnnGrasp for an example of `Roll Paper Tube' as shown in Fig.~\ref{vision} (b).

Another reason for the occurrence of the wrong prediction for grasping gesture may be that, as shown in Fig.~\ref{vision} (b), the 3d information losing problem in BOC is more serious compared with WWC. For example, in the prediction for `Can' and `Pitcher', CnnGrasp mistakenly recognized these two objects as tiny things and thus made a wrong prediction for the grasping gesture. This is common in traditional grasp classification methods due to size information lost in RGB images. In addition, from the cases of `Block' and `Food Box', it can be seen that CnnGrasp is sensitive to deformation, which is caused by taking a photo of the object from different angles.

          Although the unseen problem is a challenge in grasp pattern recognition, DcnnGrasp still obtained excellent results (higher than 94\%) on the RGB-D Object dataset and Hit-GPRec dataset. Also, it outperformed the state-of-art methods on all datasets significantly in all evaluation metrics. All results verified the strong generalizability of DcnnGrasp in unseen objects, which is attributed to the proposed training strategy based on JCEAR.



       \subsection{Robustness on the dataset with missing grasp labels}

    In the experiments given in Section \ref{sec_weak}, the robustness of DcnnGrasp was investigated on the dataset with missing grasp labels. The experimental results indicated the effectiveness of our method even when the proportion of the remaining grasp labels is small, especially on the RGB-D Object dataset and Hit-GPRec dataset. The comparision in OCS showed the strong robustness of DcnnGrasp to missing grasp labels compared with CnnGrasp. On Hit-GPRec dataset, for DcnnGrasp, the difference caused by different $n$ were less than  2\%, but for CnnGrasp, the difference had achieved 10\% to 15\% for different  valuation metrics.  Even only one object with gesture
labels in each object category appeared in the training process, the GA values of DcnnGrasp were higher than 90\% and 95\% on the RGB-D Object dataset and Hit-GPRec
dataset  respectively.
It is because DcnnGrasp can  learn and utilize the relationship
between object categories and grasping gestures well when there are sufficient   objects  with grasp labels  in each object
category appears in the training process.

    \section{Conclusion and Future Work}\label{section:conclusion and future work}
	This paper proposes a novel dual-branch convolutional neural network (DcnnGrasp) that utilizes the object category information to improve grasp pattern recognition. Meanwhile, a novel loss function JCEAR and a new training strategy are given to maximize the collaborative learning of object category classification and grasp pattern recognition. To train DcnnGrasp, two dual-label datasets are constructed based on the existing household datasets including the RGB-D Object dataset and Hig-GPRec dataset. Experimental results demonstrated the excellent performance of the proposed method.

    As a newly developed technology, computer vision-based grasp pattern recognition still has a lot of challenges. For example, how to predict human grasp affordances for a single RGB image of a scene with an arbitrary number of objects? Also, it is interesting to consider personal habits in grasp pattern recognition.

\begin{small}

\end{small}
	

\begin{thebibliography}{10}
\providecommand{\url}[1]{#1}
\csname url@samestyle\endcsname
\providecommand{\newblock}{\relax}
\providecommand{\bibinfo}[2]{#2}
\providecommand{\BIBentrySTDinterwordspacing}{\spaceskip=0pt\relax}
\providecommand{\BIBentryALTinterwordstretchfactor}{4}
\providecommand{\BIBentryALTinterwordspacing}{\spaceskip=\fontdimen2\font plus
\BIBentryALTinterwordstretchfactor\fontdimen3\font minus
  \fontdimen4\font\relax}
\providecommand{\BIBforeignlanguage}[2]{{%
\expandafter\ifx\csname l@#1\endcsname\relax
\typeout{** WARNING: IEEEtran.bst: No hyphenation pattern has been}%
\typeout{** loaded for the language `#1'. Using the pattern for}%
\typeout{** the default language instead.}%
\else
\language=\csname l@#1\endcsname
\fi
#2}}
\providecommand{\BIBdecl}{\relax}
\BIBdecl

\bibitem{wang2017robust}
Z.~Wang, J.~Merel, S.~Reed, G.~Wayne, N.~de~Freitas, and N.~Heess, ``Robust
  imitation of diverse behaviors,'' \emph{arXiv preprint arXiv:1707.02747},
  2017.

\bibitem{liu2018imitation}
Y.~Liu, A.~Gupta, P.~Abbeel, and S.~Levine, ``Imitation from observation:
  Learning to imitate behaviors from raw video via context translation,'' In
  \emph{ICRA},
  pp. 1118--1125, 2018.

\bibitem{di2020safari}
N.~Di~Palo and E.~Johns, ``Safari: Safe and active robot imitation learning
  with imagination,'' \emph{arXiv preprint arXiv:2011.09586}, 2020.

\bibitem{jin2020smart}
H.~Jin, E.~Dong, M.~Xu, and J.~Yang, ``A smart and hybrid composite finger with
  biomimetic tapping motion for soft prosthetic hand,'' \emph{JBE}, vol.~17, pp. 484--500, 2020.

\bibitem{wang2021smarthand}
X.~Wang, F.~Geiger, V.~Niculescu, M.~Magno, and L.~Benini, ``Smarthand: Towards
  embedded smart hands for prosthetic and robotic applications,'' \emph{arXiv
  preprint arXiv:2107.14598}, 2021.

\bibitem{corona2020ganhand}
E.~Corona, A.~Pumarola, G.~Alenya, F.~Moreno-Noguer, and G.~Rogez, ``Ganhand:
  Predicting human grasp affordances in multi-object scenes,'' In
  \emph{CVPR}, pp.
  5031--5041, 2020.

\bibitem{zandigohar2021multimodal}
M.~Zandigohar, M.~Han, M.~Sharif, S.~Y. Gunay, M.~P. Furmanek, M.~Yarossi,
  P.~Bonato, C.~Onal, T.~Padir, D.~Erdogmus \emph{et~al.}, ``Multimodal fusion
  of emg and vision for human grasp intent inference in prosthetic hand
  control,'' \emph{arXiv preprint arXiv:2104.03893}, 2021.

\bibitem{ghazaei2019grasp}
G.~Ghazaei, F.~Tombari, N.~Navab, and K.~Nazarpour, ``Grasp type estimation for
  myoelectric prostheses using point cloud feature learning,'' \emph{arXiv
  preprint arXiv:1908.02564}, 2019.

\bibitem{zandigohar2021netcut}
M.~Zandigohar, D.~Erdogmus, and G.~Schirner, ``Netcut: Real-time dnn inference
  using layer removal,'' \emph{arXiv preprint arXiv:2101.05363}, 2021.

\bibitem{veres2017modeling}
M.~Veres, M.~Moussa, and G.~W. Taylor, ``Modeling grasp motor imagery through
  deep conditional generative models,''  \emph{RAL}, vol.~2, no.~2, pp. 757--764, 2017.

\bibitem{weiner2018kit}
P.~Weiner, J.~Starke, F.~Hundhausen, J.~Beil, and T.~Asfour, ``The kit
  prosthetic hand: design and control,'' In \emph{IROS}, pp. 3328--3334, 2018.

\bibitem{liu2021novel}
S.~Liu, M.~Van, Z.~Chen, J.~Angeles, and C.~Chen, ``A novel prosthetic finger
  design with high load-carrying capacity,'' \emph{Mechanism and Machine
  Theory}, vol. 156, p. 104121, 2021.

\bibitem{yusof2021design}
K.~H. Yusof, M.~A. Zulkipli, A.~S. Ahmad, M.~F. Yusri, S.~Al-Zubaidi, and
  M.~Mohammed, ``Design and development of prosthetic leg with a mechanical
  system,'' In \emph{ICSGRC}, pp. 217--221, 2021.

\bibitem{kumra2017robotic}
S.~Kumra and C.~Kanan, ``Robotic grasp detection using deep convolutional
  neural networks,'' In \emph{IROS}, pp. 769--776, 2017.

\bibitem{du2019vision}
G.~Du, K.~Wang, and S.~Lian, ``Vision-based robotic grasping from object
  localization pose estimation grasp detection to motion planning: A review,''
  \emph{arXiv preprint arXiv:1905.06658}, 2019.

\bibitem{caldera2018review}
S.~Caldera, A.~Rassau, and D.~Chai, ``Review of deep learning methods in
  robotic grasp detection,'' \emph{Multimodal Technologies and Interaction},
  vol.~2, no.~3, p.~57, 2018.

\bibitem{song2020novel}
Y.~Song, L.~Gao, X.~Li, and W.~Shen, ``A novel robotic grasp detection method
  based on region proposal networks,'' \emph{Robotics and Computer-Integrated
  Manufacturing}, vol.~65, p. 101963, 2020.

\bibitem{deng2019attention}
Z.~Deng, G.~Gao, S.~Frintrop, F.~Sun, C.~Zhang, and J.~Zhang, ``Attention based
  visual analysis for fast grasp planning with a multi-fingered robotic hand,''
  \emph{Frontiers in neurorobotics}, vol.~13, p.~60, 2019.

\bibitem{taverne2019video}
L.~T. Taverne, M.~Cognolato, T.~B{\"u}tzer, R.~Gassert, and O.~Hilliges,
  ``Video-based prediction of hand-grasp preshaping with application to
  prosthesis control,'' In \emph{ICRA}, pp. 4975--4982, 2019.

\bibitem{hundhausen2019resource}
F.~Hundhausen, D.~Megerle, and T.~Asfour, ``Resource-aware object
  classification and segmentation for semi-autonomous grasping with prosthetic
  hands,'' In \emph{2019 IEEE-RAS 19th International Conference on Humanoid
  Robots}, pp. 215--221, 2019.

\bibitem{kara2019modeling}
T.~Kara and A.~S. Masri, ``Modeling and analysis of a visual feedback system to
  support efficient object grasping of an emg-controlled prosthetic hand,''
  \emph{CDBME}, vol.~5, no.~1, pp.
  207--210, 2019.

\bibitem{gu2018recent}
J.~Gu, Z.~Wang, J.~Kuen, L.~Ma, A.~Shahroudy, B.~Shuai, T.~Liu, X.~Wang,
  G.~Wang, J.~Cai \emph{et~al.}, ``Recent advances in convolutional neural
  networks,'' \emph{Pattern Recognition}, vol.~77, pp. 354--377, 2018.

\bibitem{krizhevsky2012imagenet}
A.~Krizhevsky, I.~Sutskever, and G.~E. Hinton, ``Imagenet classification with
  deep convolutional neural networks,''In \emph{NIPS}, vol.~25, pp. 1097--1105, 2012.

\bibitem{RN2}
G.~Ghazaei, A.~Alameer, P.~Degenaar, G.~Morgan, and K.~Nazarpour, ``Deep
  learning-based artificial vision for grasp classification in myoelectric
  hands,''\emph{JNE}, vol.~14, no.~3, p. 036025,
  2017.

\bibitem{bertsekas1997nonlinear}
D.~P. Bertsekas, ``Nonlinear programming,'' \emph{JORS}, vol.~48, no.~3, pp. 334--334, 1997.

\bibitem{dovsen2011transradial}
S.~Do{\v{s}}en and D.~B. Popovi{\'c}, ``Transradial prosthesis: artificial
  vision for control of prehension,'' \emph{Artificial organs}, vol.~35, no.~1,
  pp. 37--48, 2011.

\bibitem{wake2021object}
N.~Wake, D.~Saito, K.~Sasabuchi, H.~Koike, and K.~Ikeuchi, ``Object affordance
  as a guide for grasp-type recognition,'' \emph{arXiv preprint
  arXiv:2103.00268}, 2021.

\bibitem{kopicki2016one}
M.~Kopicki, R.~Detry, M.~Adjigble, R.~Stolkin, A.~Leonardis, and J.~L. Wyatt,
  ``One-shot learning and generation of dexterous grasps for novel objects,''
  \emph{IJRR}, vol.~35, no.~8, pp.
  959--976, 2016.

\bibitem{shi2020computer}
C.~Shi, D.~Yang, J.~Zhao, and H.~Liu, ``Computer vision-based grasp pattern
  recognition with application to myoelectric control of dexterous hand
  prosthesis,'' \emph{IEEE Transactions on Neural Systems and Rehabilitation
  Engineering}, vol.~28, no.~9, pp. 2090--2099, 2020.

\bibitem{han2019hand}
M.~Han, S.~Y. G{\"u}nay, {\.I}.~Yildiz, P.~Bonato, C.~D. Onal, T.~Padir,
  G.~Schirner, and D.~Erdo{\u{g}}mu{\c{s}}, ``From hand-perspective visual
  information to grasp type probabilities: deep learning via ranking labels,''
  In \emph{12th ACM international conference on pervasive technologies related
  to assistive environments}, pp. 256--263, 2019.

\bibitem{zandigohar2019towards}
M.~Zandigohar, M.~Han, D.~Erdo{\u{g}}mu{\c{s}}, and G.~Schirner, ``Towards
  creating a deployable grasp type probability estimator for a prosthetic
  hand,''  \emph{Cyber Physical Systems. Model-Based Design}, pp. 44--58,
  2019.

\bibitem{lai2011large}
K.~Lai, L.~Bo, X.~Ren, and D.~Fox, ``A large-scale hierarchical multi-view
  rgb-d object dataset,'' In \emph{ICRA}, pp. 1817--1824, 2011.

\bibitem{geusebroek2005amsterdam}
J.-M. Geusebroek, G.~J. Burghouts, and A.~W. Smeulders, ``The amsterdam library
  of object images,'' \emph{IJCV}, vol.~61,
  no.~1, pp. 103--112, 2005.

\bibitem{nene1996columbia}
S.~A. Nene, S.~K. Nayar, H.~Murase \emph{et~al.}, ``Columbia object image
  library (coil-100),'' 1996.

\bibitem{cognolato2020gaze}
M.~Cognolato, A.~Gijsberts, V.~Gregori, G.~Saetta, K.~Giacomino, A.-G.~M.
  Hager, A.~Gigli, D.~Faccio, C.~Tiengo, F.~Bassetto \emph{et~al.}, ``Gaze,
  visual, myoelectric, and inertial data of grasps for intelligent
  prosthetics,'' \emph{Scientific data}, vol.~7, no.~1, pp. 1--15, 2020.

\bibitem{huang2017densely}
G.~Huang, Z.~Liu, L.~Van Der~Maaten, and K.~Q. Weinberger, ``Densely connected
  convolutional networks,'' In \emph{CVPR}, pp. 4700--4708, 2017.

\bibitem{deng2009imagenet}
J.~Deng, W.~Dong, R.~Socher, L.-J. Li, K.~Li, and L.~Fei-Fei, ``Imagenet: A
  large-scale hierarchical image database,'' In \emph{CVPR}, pp. 248--255, 2009.

\bibitem{tan2019efficientnet}
M.~Tan and Q.~Le, ``Efficientnet: Rethinking model scaling for convolutional
  neural networks,'' In \emph{ICML},
  pp. 6105--6114, 2019.

\bibitem{tan2021efficientnetv2}
M.~Tan and Q.~V. Le, ``Efficientnetv2: Smaller models and faster training,''
  \emph{arXiv preprint arXiv:2104.00298}, 2021.

\bibitem{jha2020lightlayers}
D.~Jha, A.~Yazidi, M.~A. Riegler, D.~Johansen, H.~D. Johansen, and
  P.~Halvorsen, ``Lightlayers: Parameter efficient dense and convolutional
  layers for image classification,'' In \emph{PDCAT}, pp.
  285--296, 2020.

\bibitem{han2020ghostnet}
K.~Han, Y.~Wang, Q.~Tian, J.~Guo, C.~Xu, and C.~Xu, ``Ghostnet: More features
  from cheap operations,'' In \emph{CVPR}, pp. 1580--1589, 2020.

\bibitem{radosavovic2020designing}
I.~Radosavovic, R.~P. Kosaraju, R.~Girshick, K.~He, and P.~Doll{\'a}r,
  ``Designing network design spaces,'' In \emph{CVPR}, pp. 10\,428--10\,436, 2020.

\bibitem{cutkosky1989grasp}
M.~R. Cutkosky \emph{et~al.}, ``On grasp choice, grasp models, and the design
  of hands for manufacturing tasks.'' \emph{IEEE Transactions on robotics and
  automation}, vol.~5, no.~3, pp. 269--279, 1989.

\bibitem{bullock2013grasp}
I.~M. Bullock, J.~Z. Zheng, S.~De~La~Rosa, C.~Guertler, and A.~M. Dollar,
  ``Grasp frequency and usage in daily household and machine shop tasks,''
  \emph{ToH}, vol.~6, no.~3, pp. 296--308, 2013.

\bibitem{robert2014machine}
C.~Robert, ``Machine learning, a probabilistic perspective,'' 2014.

\bibitem{thabtah2009naive}
F.~Thabtah, M.~Eljinini, M.~Zamzeer, and W.~Hadi, ``Na{\"\i}ve bayesian based
  on chi square to categorize arabic data,'' In \emph{11th
  international business information management association conference (IBIMA)
  conference on innovation and knowledge management in twin track economies,
  Cairo, Egypt}, pp. 4--6, 2009.

\bibitem{opitz2019macro}
J.~Opitz and S.~Burst, ``Macro f1 and macro f1,'' \emph{arXiv preprint
  arXiv:1911.03347}, 2019.

\bibitem{zhang2018deep}
Y.~Zhang, T.~Xiang, T.~M. Hospedales, and H.~Lu, ``Deep mutual learning,'' In
  \emph{CVPR}, pp. 4320--4328, 2018.
\end{thebibliography}
\end{document}